\documentclass[sn-basic,iicol]{sn-jnl}

\jyear{2022}%

\theoremstyle{thmstyleone}%
\theoremstyle{thmstyletwo}%

\theoremstyle{thmstylethree}%

\usepackage{algorithm}
\usepackage{color}

\usepackage{graphicx}
\usepackage{amsmath}
\usepackage{amssymb}
\usepackage{booktabs}
\usepackage{amsfonts}

\usepackage{bbding}
\usepackage{multirow}
\usepackage{color}

\begin{document}

\title[Article Title]{PartFormer: Awakening Latent Diverse Representation from Vision Transformer for Object Re-Identification}


\author[1]{\fnm{Lei} \sur{Tan}}\email{tanlei@stu.xmu.edu.cn}

\author[1]{\fnm{Pingyang} \sur{Dai}}\email{pydai@xmu.edu.cn}

\author[3,4]{\fnm{Jie} \sur{Chen}}\email{chenj@pcl.ac.cn}

\author[1]{\fnm{Liujuan} \sur{Cao}}\email{caoliujuan@xmu.edu.cn}

\author[5]{\fnm{Yongjian} \sur{Wu}}\email{littlekenwu@tencent.com}

\author[1,2]{\fnm{Rongrong} \sur{Ji}}\email{rrji@xmu.edu.cn}

\affil[1]{\orgdiv{Key Laboratory of Multimedia Trusted Perception and Efficient Computing, Ministry of Education of China}, \orgname{Xiamen University}, \orgaddress{\city{Xiamen}, \country{China}}}

\affil[2]{\orgdiv{Institute of Artificial Intelligence, Xiamen University}, \orgname{Xiamen University}, \orgaddress{\city{Xiamen},  \country{China}}}

\affil[3]{\orgdiv{School of Electronic and Computer Engineering}, \orgname{Peking University}, \orgaddress{\city{Shenzhen},  \country{China}}}

\affil[4]{\orgname{Peng Cheng Laboratory}, \orgaddress{\city{Shenzhen},  \country{China}}}

\affil[5]{\orgdiv{Youtu Lab}, \orgname{Tencent}, \orgaddress{\city{Shanghai},  \country{China}}}


\abstract{
Extracting robust feature representation is critical for object re-identification to accurately identify objects across non-overlapping cameras. Although having a strong representation ability, the Vision Transformer (ViT) tends to overfit on most distinct regions of training data, limiting its generalizability and attention to holistic object features. Meanwhile, due to the structural difference between CNN and ViT, fine-grained strategies that effectively address this issue in CNN do not continue to be successful in ViT. To address this issue, by observing the latent diverse representation hidden behind the multi-head attention, we present PartFormer, an innovative adaptation of ViT designed to overcome the granularity limitations in object Re-ID tasks. The PartFormer integrates a Head Disentangling Block (HDB) that awakens the diverse representation of multi-head self-attention without the typical loss of feature richness induced by concatenation and FFN layers post-attention. To avoid the homogenization of attention heads and promote robust part-based feature learning, two head diversity constraints are imposed: attention diversity constraint and correlation diversity constraint. These constraints enable the model to exploit diverse and discriminative feature representations from different attention heads. Comprehensive experiments on various object Re-ID benchmarks demonstrate the superiority of the PartFormer. Specifically, our framework significantly outperforms state-of-the-art by 2.4\% mAP scores on the most challenging MSMT17 dataset.
}




\maketitle

\section{Introduction}

Object re-identification (Re-ID) which involves retrieving a specific object across a distributed set of non-overlapping cameras, has attracted significant interest from academia to industry due to its crucial role in surveillance systems~\cite{he2021transreid,wang2022pose,tan2022dynamic}. As deep learning took over, particularly after the inception of part-based models, this area rapidly blossomed. Various variations of the part-based model demonstrate that employing part-level features for object image description offers fine-grained information and is greatly beneficial for object Re-ID in CNN-based methods~\cite{sun2018beyond,zhu2020identity,li2021diverse}.

\begin{figure}[t]
\centering
\includegraphics[width=\columnwidth]{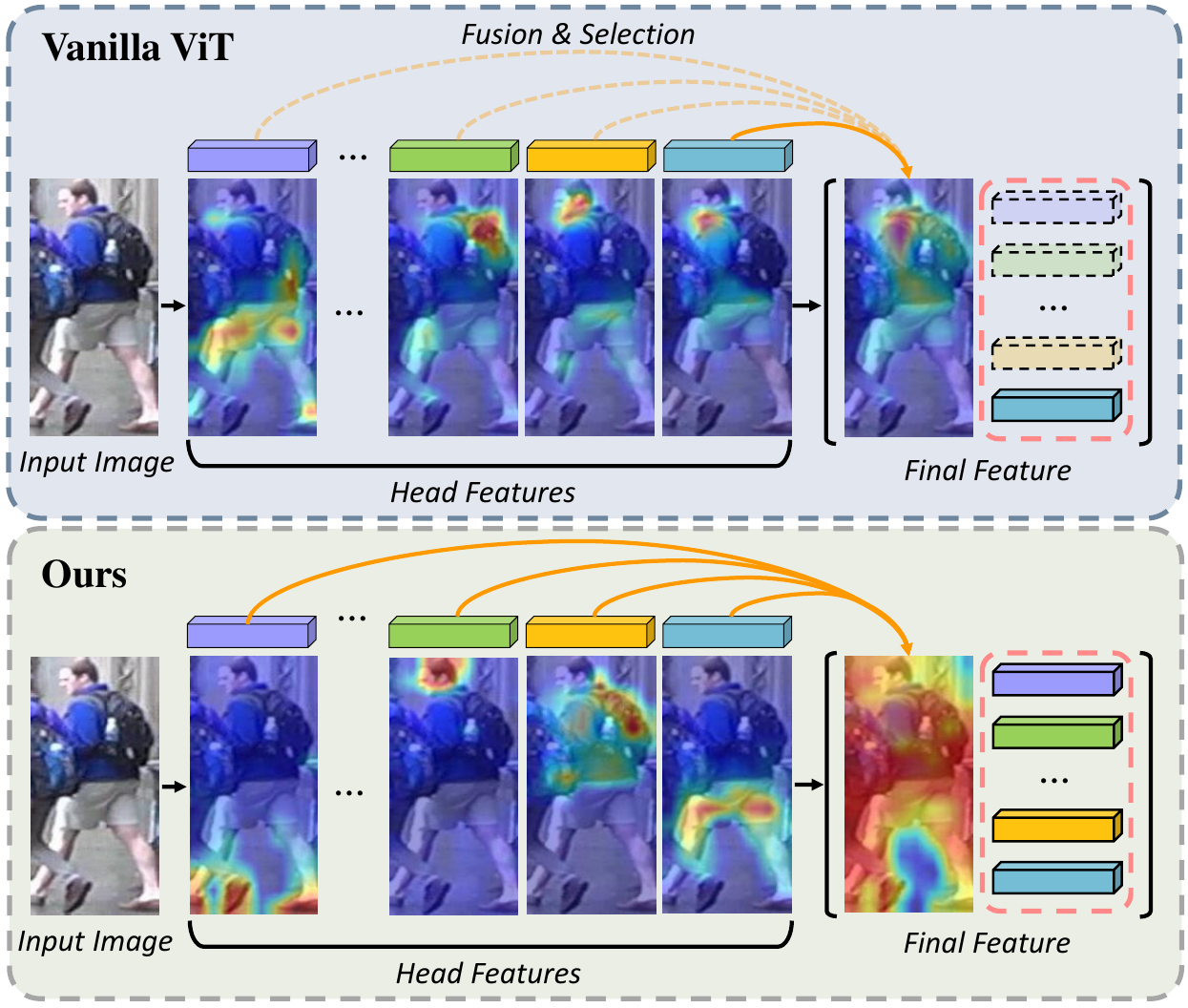}
\caption{\textbf{The motivation of proposed PartFomer.} We visualize the head's attention in the final ViT block to explain the advantage of our method. \textbf{In vanilla ViT (Top)}, we observe that multi-head attention contains a diverse focus on different parts of the object. But, after the subsequent layers, most of them are lost. This yields incomplete representation and limits the re-identification performance. \textbf{In our proposed Partformer (Bottom)}, we discover the cause of such degradation and awaken the latent diverse representation behind the multi-head attention.}
  \label{fig:motivation}
\end{figure}

Looking back at recent years, despite ongoing improvements in the CNN-based part-based models, the revolution of Vision Transformer swept across the landscape of computer vision. The global receptive capabilities of multi-head self-attention in ViT have demonstrated a robust capacity for extracting distinguishing features, and advancing areas such as classification, detection, segmentation, and Re-ID. The impressive power of ViT easily pushes the boundary of the object Re-ID topic. Therefore, it is intuitive to insert the part-based descriptions in a ViT model to improve its ability in object re-identification. Influenced by the success of CNN-based part models, recent works still attempt to achieve part-level feature representation by adding extra fine-grained modules~\cite{wang2022pose} or segmenting the image tokens~\cite{he2021transreid} in ViT architecture. However, as shown at the top of Fig.~\ref{fig:motivation}, an interesting observation of ViT in the Re-ID task is that during the multi-head self-attention layer, different heads can display a diverse focus on the different human parts. However, these diverse attentions are not effectively translated into the final representation. As shown at the top of Fig.~\ref{fig:motivation}, although the multi-head attention may show the attention on the diverse parts, many valuable part representations are lost after the combination of concatenation, linear projection, and FFN layer. This combination acts like a gate to pick the most discriminative features from fusion features, while fine-grained features with less discriminative will be neglected.

This observation led us to realize that the ViT architecture inherently has diverse fine-grained representations. \textbf{Rather than incrementally modifying the original architecture as current practices suggest~\cite{he2021transreid,wang2022pose}, it is more effective to eliminate structural components that hinder the diversity of these fine-grained representations.} 
Hence, from this perspective, we thoroughly analyzed the latent diversity representation in the ViT structure, leading to the discovery of the negative impact of the combination of concatenation, linear projection, and FFN layer in the multi-head self-attention module on these representations.
Thus, based on the analyses, we propose the PartFormer, a novel part-based vision transformer to awaken the latent diverse representation hidden in the multi-heads. The PartFormer integrates a Head Disentangling Block (HDB), which preserves the inherent diverse attention hidden in the multi-heads but discards the concatenation layer, retaining broader usable fine-grained features for Re-ID tasks. Though the diversity inherited from the pertained model makes sense in training, without an explicit constraint, inevitable degradation can easily push different heads' representations to become more similar to each other. Thus, we incorporate head diversity constraints to guide the learning of different heads. These constraints are twofold, including an attention diversity constraint and a correlation diversity constraint.  The former attention diversity constraint is widely used in previous works~\cite{tan2022dynamic} to encourage distinct heads to focus on different regions of the same input image, optimizing the dissimilarity between the attention results they extract. The correlation diversity constraint considers the representation diverse from the dataset level, which aims to push the final similarity for a person in different heads to show different correlation distributions with other people's. More specifically, the correlation diversity constraint encourages two samples that are similar in one head to be represented differently in other heads. As shown at the bottom of Fig.~\ref{fig:motivation}, by combining the Head Disentangling Block and Head Diversity Constraints, the proposed PartFormer can leverage part-based fine-grained representation to achieve a more efficient object Re-ID.

The main contributions of this paper can be summarized as follows:
\begin{itemize}
    \item We investigate the limitation of the vanilla ViT in fine-grained representation and further introduce a novel Partformer network that combines the part-based conception within a vision transformer framework to overcome the granularity limitations in the object Re-ID.
    
    \item We unveil the latent diverse representation within the multi-head attention mechanism and awaken it by proposing a Head Disentangling Block alongside Head Diversity Constraints to maximize the diversity of heads' representation in the PartFormer.
    
    \item Extensive experiments on public object re-identification datasets, such as MSMT17, Market1501, DukeMTMC, Occluded-Duke, VeRi-776, and VehicleID, validate the superior performance of our proposed PartFormer.
\end{itemize}

The remainder of this paper is organized as follows:
Section~\ref{sec:RelateWork} presents a brief literature review of the related work.
The proposed adversarial Partformer model for object re-identification is described in detail in Section~\ref{sec:Method}.
The configurations and results of experiments are presented in Section~\ref{sec:Experiments}.
Finally, the conclusion of this paper is summarized in Section~\ref{sec:Conclusion}.


\section{Related Works}
\label{sec:RelateWork}
The studies of object ReID have been mainly focused on two typical topics: person ReID and vehicle ReID. For both re-identification tasks, the methods can be roughly divided into the global-based method or part-based method.

\textbf{Person re-identification} addresses the problem of matching pedestrian images across disjoint cameras.
Zheng \emph{et al}.~\cite{zheng2019joint} integrate discriminative and generative learning in a single unified network for person re-identification.
Ye \emph{et al}.~\cite{zheng2019joint} propose the AGW, which combines the non-local attention block, generalized-mean pooling, and weighted regularization triplet together in a unified framework.
Tan \emph{et al}.~\cite{tan2022dynamic} proposes a dynamic prototype mask to align the occluded person in a learnable way.
Zhang \emph{et al}.~\cite{zhang2023pha} finds the weakness of vit in capturing high-frequency components and proposes patch-wise high-frequency augmentation to extract discriminative person representations.
Besides using the global feature representation~\cite{luo2019bag,zheng2019joint,ye2021deep}, employing part-based features to offer fine-grained information is also a mainstream strategy that has been verified as beneficial for person ReID~\cite{kim2023partmix,ni2023part,zhu2022pass}.
Methods like PCB~\cite{sun2018beyond}, MGN~\cite{wang2018learning}, and Pyramid~\cite{zheng2019pyramidal} horizontally divide the input images or feature maps into several parts to conduct a fine-grained representation. Several works also attempted to spread the part model in the ViT. TransReID~\cite{he2021transreid} firstly takes advantage of the ViT structure and proposes a jigsaw patches module to obtain perturbation-invariant representation. By employing the key point results provided by the extra pose estimation network, PFD~\cite{wang2022pose} proposes a pose-guided feature disentangling method for occluded person re-identification. Although several works also highlight the importance of fine-grained representation in vision transformers, they either highly rely on the prior segmentation results provided by pre-trained networks or lack efficient design to obtain rich and diverse fine-grained representation.

\textbf{Vehicle re-identification} aims to locate and recognize a vehicle of interest across multiple non-overlapping cameras in various traffic intersections~\cite{zhao2021phd}. 
Combining the global and local features is also a widely used strategy to obtain robust and discriminative representations.  PVEN~\cite{PVEN} introduced a parsing-based view-aware embedding network to achieve the view-aware feature alignment. He \emph{et al}.~\cite{he2019part} develop a novel framework to integrate part constraints with the global Re-ID modules by introducing a detection branch. PCR-Net~\cite{liu2020beyond} builds a part-neighboring graph to explicitly model the correlation among parts, which can discover the most effective local features of varied viewpoints. GiT~\cite{shen2023git} couples graph networks with vision transformers, bringing effective cooperation between local and global features. Benefiting from the detailed parsing and keypoint annotations on the vehicle re-identification datasets, the vehicle re-identification method can easily obtain a precise perception of different parts of a vehicle. Since PartFormer follows the setting of TransReID~\cite{he2021transreid} which only contains the identity label and viewpoint label, it may cause an unfair comparison between PartFormer with those methods that use more comprehensive pixel-level annotations. Nonetheless, PartFormer still shows a strong ability in this topic with a more limited training setting.

\section{Method}
\label{sec:Method}
\subsection{Overall Framework}
\begin{figure*}[t]
\centering
\includegraphics[width=2.1\columnwidth]{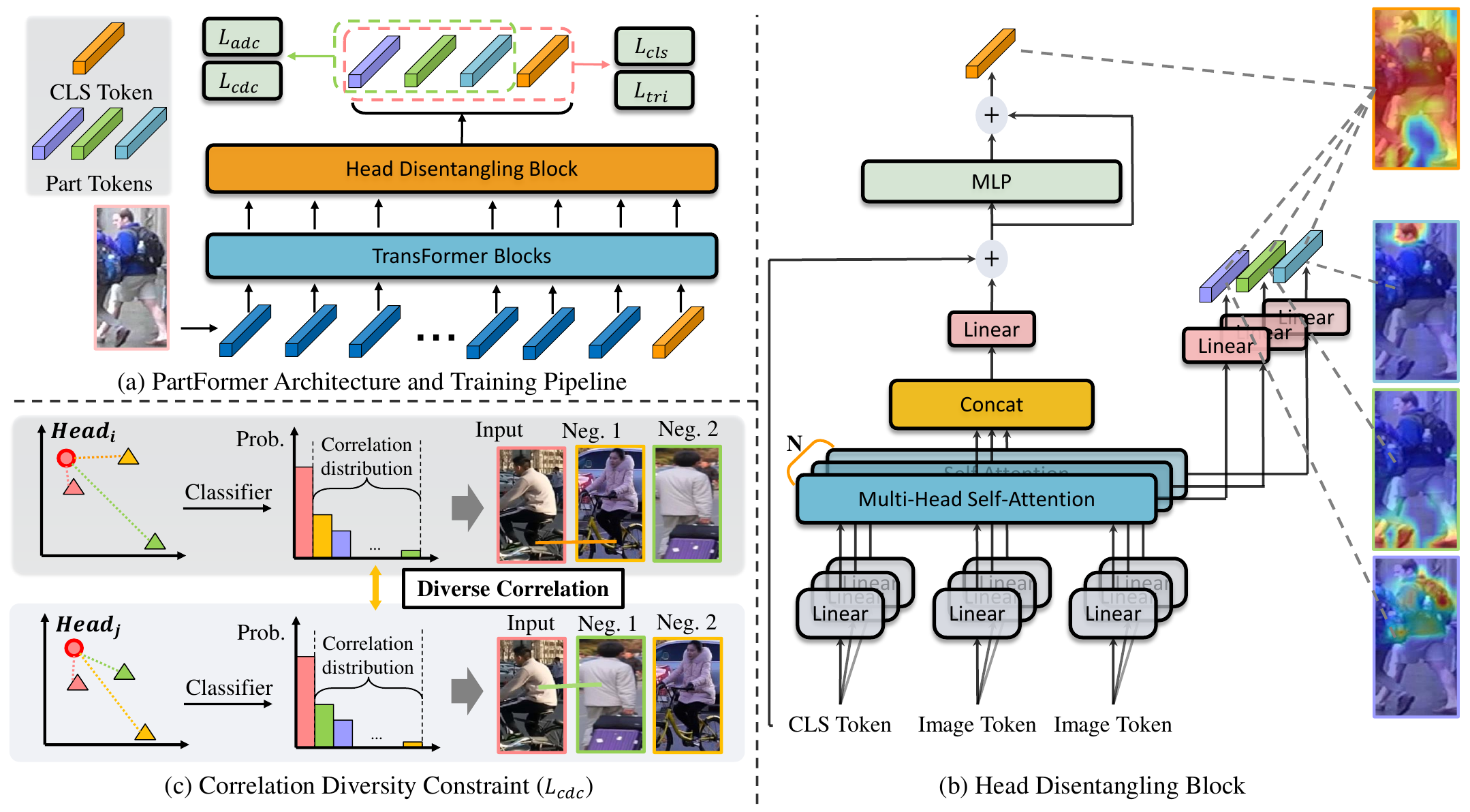}
\caption{\textbf{(a) The Architecture and Training Pipeline of PartFormer.} PartFromer constructs by basic Transformer blocks inherited from the vanilla ViT, while in the last block, the head disentangling block takes the place of the original transformer block to awaken the latent diverse representation hidden behind the multiple heads. \textbf{(b) The Architecture of Head Disentangling Block (HDB).} HDB aims to get rid of the fusion and selection processing in the transformer block. In HDB, after the multi-head attention operation, the heads' diverse representation is directly output with an unshared linear projection. \textbf{(c) The Explanation of Correlation Diversity Constraint ($\mathcal{L}_{cdc}$).} The $\mathcal{L}_{cdc}$ considers that after removing the score obtained by ground truth, the distribution after the classifier is a correlation distribution for each input image. $\mathcal{L}_{cdc}$ encourages the different heads to show a different correlation distribution for the same input image. Here, we show an example of $\mathcal{L}_{cdc}$ after training in the MSMT17. For the input image Input, Neg.~1 and Neg.~2 are the two hardest negative samples in $head_i$ and $head_j$ respectively. Due to the $\mathcal{L}_{cdc}$, the similarity on the bike in $head_i$ will prevent Input and Neg.~1 from still similar in $head_j$, thus pushing $head_j$ to focus other content such as clothing.}
\label{fig:overview}
\end{figure*}
Based on the observations made above, we propose PartFormer, an efficient framework for object Re-ID. The overview of our proposed PartFormer framework is illustrated in Fig.~\ref{fig:overview}. The PartFormer adopts a pre-trained vanilla ViT~\cite{dosovitskiy2020image} as its basic backbone network. Herein, we denote the input image as $I$ with the resolution as $H\times W$. We first split the image into $M (h \times w)$ patches with the size $P$. Specifically, it can be described as:
\begin{equation}
M = h \times w = \left \lfloor \frac{H+s-P}{s} + \frac{W+s-P}{s} \right \rfloor,
\end{equation}
where the $s$ refer to the step size of the sliding window. After the linear projection $\mathcal{F}$, a learnable class token $x_{cls}$ is attached to aggregate the information from image patches. Following the TransReID~\cite{he2021transreid}, a learnable position embedding $\mathcal{P}$ and side information embedding $\mathcal{S}$ are added to preserve positional information and side information respectively, which can be formulated as:
\begin{equation}
z_0 = [x^0_{cls};\mathcal{F} (x^0_1);\mathcal{F}(x^0_2);\cdots;\mathcal{F}(x^0_K)] + \mathcal{P} + \lambda \mathcal{S},
\end{equation}
where $z_0$ is the input of the transformer blocks, and $x^0_k$ refers to the $k_{th}$ patch feature after patchify operation. The hyper-parameter $\lambda$ is used to balance the weight of side information embedding. Following the setting of TransReID~\cite{he2021transreid}, we set the $\lambda = 3.0$ in the whole experiment.

To encode the input $z_0$ into a latent representation space, we carry out a vision transformer backbone consisting of $L$ blocks. During this process, we replaced the final transformer block in the Vision Transformer with the Head Disentangling Block (HDB) to encourage the entire network to provide fine-grained feature representations. To keep the diversity of different heads in the HDB, head diversity constraints which include an attention diversity constraint and a correlation diversity constraint, are introduced as well.

\subsection{Head Disentangling Block}
Before introducing the structure of the Head Disentangling Block, it is necessary to look back to the observations shown at the top of Fig.~\ref{fig:motivation}. Although the multi-head self-attention layer (MHSA) has already brought a level of diversity to the feature representations, this kind of diversity is not well transmitted to the final representation, which limits the performance of ViT in the object Re-ID.

To explore the problem behind it, we start with the basic structure of a transformer block, which consists of a multi-head self-attention layer and a feed-forward network (FFN). In particular, a single-head attention for the $i_{th}$ head in the $l_{th}$ block is computed as below: 
\begin{equation}
\begin{split}
\label{eq:attn}
&Attn(Q_{i,l},K_{i,l},V_{i,l})=a_{i,l}V_{i,l},\\
&with \qquad a_{i,l} = \sigma(\frac{Q_{i,l}(K_{i,l})^T}{\sqrt{d}}), \\
&and  \qquad  (Q_{i,l},K_{i,l},V_{i,l}) = (z_lW_{i,l}^{Q}, z_lW_{i,l}^{K}, z_lW_{i,l}^{V}),
\end{split}
\end{equation}
where $Q_{i,l}$, $K_{i,l}$, and $V_{i,l}$ are query, key, and value matrices, while $W_{i,l}^{Q}$, $W_{i,l}^{K}$, $W_{i,l}^{V}$ are the parameter matrices in the $i_{th}$ attention head of the $l_{th}$ transformer block, respectively. 
$\sigma(\cdot)$ refers to the softmax function, $d$ is a scaling factor, and $z_l$ indicates the input for the $l_{th}$ block. 
To effectively aggregate attention results on different head representations, the vanilla ViT directly concatenates the outputs from different heads and projects them with a parameter matrix as follows:
\begin{equation}
MSA(z_l)=Concat(a_{0,l}V_{0,l},\dots  ,a_{N,l}V_{N,l})W_{l},
\label{eq:msa}
\end{equation}
where the $W_l \in \mathbb{R}^{c\times c}$ refers to the weight of linear projection in the $i_{th}$ attention head of the $l_{th}$ transformer block, and $N$ is the number of heads.
With the FFN and residual connections, the final output of a transformer block is:
\begin{equation}
z_{l}^{msa} = z_{l} + MSA(z_{l}), z_{l+1}=z_{l}^{msa}+FFN(z_{l}^{msa}).
\label{eq:ffn}
\end{equation}
Based on Eq.~(\ref{eq:msa}), if we consider the equation from the head's side, it can also be written as:
\begin{equation}
MSA(z_l)=a_{0,l}V_{0,l}W_{0,l}+\dots+a_{N,l}V_{N,l}W_{N,l},
\label{eq:msa2}
\end{equation}
where the $W_{i,l} \in \mathbb{R}^{\frac{c}{N} \times c}$ is submatrix divided from the $W_l$.

From Eq.~(\ref{eq:msa2}), we can observe that besides the attention results, which are shown in Fig.~\ref{fig:motivation}, the aggregation output of the MHSA layer is also greatly influenced by two parts: value matrices $V_{i,l}$ and the linear projection $W_l$. If we consider the different weight matrix $W_{i,l}$ as a kind of gate layer to aggregate the multi-head representations, it is obvious that if one of the $W_{i,l}$ weakens during training, even if its attention shows the information, the representation will still be highly ignored in the final representation. Meanwhile, to make matters worse, the $V_{i,l}$ will be neglected and will receive insufficient training. Besides, the situation is the same if we continue to divide the following FFN layer. Such a pipeline ensures that the most discriminative features can be comfortably trained and transferred, while processing is much more difficult for those fine-grained features with less discriminative. A powerful argument for this observation comes from the adaViT~\cite{meng2022adavit}, which indicates that many heads in a ViT can be sparse with limited degradation.

Therefore, based on the above observations, we attempt to awaken the latent diverse representations that are hidden behind MHSA and further propose a Head Disentangling Block (HDB) to replace the last transformer block in the vision transformer. As shown in Fig.~\ref{fig:overview}, in the HDB, we break the addition operation in Eq.~(\ref{eq:msa2}), and force each head to represent a part feature. Thus, in the output of MHSA, the representations of all heads can be considered equally and push a more complete feature representation. However, if we further consider Eq.~(\ref{eq:attn}) and Eq.~(\ref{eq:ffn}), there are still several nuisances. The first one is the residual connection, which also introduces an additional operation in Eq.~(\ref{eq:ffn}). Since all the heads share the same input $z_l$, this operation may prevent the multi-head from exhibiting diversity. Another is the attention between the class token itself, which may also prevent the class token in the HDB from extracting the features of image tokens. Therefore, in the HDB, we removed these two parts during the computation of part features. Finally, the output of the part feature $f_{i}^{part}$ can be given as:
\begin{equation}
f_{i}^{part}=a^{I}_{i,l}V_{i,l}W_{i,l},
\label{eq:hdb}
\end{equation}
where $a^{I}_{i,l}$ is the attention matrix that only considers the results between the class token and image tokens.

Furthermore, since the part features do not need to perform feature selection from aggregated multi-head representations like the original type of MHSA, it is interesting to find that we can even remove the subsequent FFN layers without performance degradation, which largely increases the efficiency of the HDB.

\subsection{Head Diversity Constraints}
Although all heads in the HDB have some diversity due to their different initial weights inherited from the pertained ViT, without explicit constraints, those representations will be inevitably homogenizing due to the same input and training target, resulting in a suboptimal solution. To tackle this issue, in PartFormer, we incorporate two constraints, including an attention diversity constraint and a correlation diversity constraint, to maintain the diversity of different heads. 

The attention diversity constraint aims to push the different heads to focus on different discriminative foregrounds. Generally, in the HDB, we consider that the feature of each head comes from the image feature aggregation guided by the attention matrix $a_{i,l}$. Therefore, to encourage each head to focus on different fine-grained patterns, the attention diversity constraint is shown as follows:
\begin{equation}
\mathcal{L}_{hdc}=\left \| a_la_l^{T} - \mathbb{I}_N  \right \|_1,
\label{eq:hdc}
\end{equation}
where $a_l\in \mathbb{R}^{N\times M}$ is the attention matrix of different heads in the HDB, while $\mathbb{I}_N \in N \times N$ is the target identity matrix. The $\left \| \cdot \right \|_1$ refers to the  1-norm distance between the attention matrix and target matrix.

Although attention diversity constraint works well in discovering different fine-grained regions, it struggles to achieve discriminative diversity. Since the position embedding is learnable, image features are also influenced by their position. Though heads are extracted from different regions, they may focus on patterns with high correlation, such as the left arm and right arm with similar discriminative. Obviously, it may not be necessary to use multi-heads to represent high correlation patterns. Thus, to address the issue, a correlation diversity loss is presented. Generally, the weight matrix of a classifier can be considered a prototype matrix, where each row represents a specific class. Therefore, the output of the classifier can be regarded as the similarity between the input image and different classes. If we exclude the input sample and its corresponding prototype, we can obtain a similarity distribution between the feature and other class features. Apparently, if two heads extract content with a high correlation, they will also show a high similarity in the similarity distribution. 

Therefore, as shown in Fig.~\ref{fig:overview} (c), the correlation diversity loss employs the similarity distribution for each sample and pushes them away from each other during training. More specifically, for the distribution $D_i$ of features $f_{i}^{part}$ obtained through its corresponding classifiers, the correlation diversity loss can be shown as:
{\small
\begin{equation}
\mathcal{L}_{cdc}=\frac{2}{N(N-1)} \sum_{N-1}^{i=1}\sum_{N}^{j=i+1}(\sigma(M(D_i,y))\sigma(M(D_j, y))^T).
\label{eq:cdc}
\end{equation}}

Here, $y$ refers to the true label of the input $f_{i}^{part}$ while $M(D, y)$ indicates the operation that excludes the similarity between the $f_{i}^{part}$ and its corresponding prototype in the distribution $D$.

\subsection{Training and Inference}
Apart from the above-mentioned head diversity constraints, the widely known identity-based softmax cross-entropy loss $\mathcal{L}_{cls}$ and triplet loss $\mathcal{L}_{tri}$ with soft-margin are also employed in the training of PartFormer. Thus, the objective function of PartFormer can be formulated as:
\begin{equation}
\begin{split}
\label{eq:loss}
&\mathcal{L} = \mathcal{L}_{g} + \mathcal{L}_{p} + \alpha \mathcal{L}_{adc} + \beta \mathcal{L}_{cdc}  \\
&with \qquad \mathcal{L}_{g} = \mathcal{L}_{cls}(f^g) + \mathcal{L}_{tri}(f^g)\\
&and  \qquad  {\mathcal{L}}_{p} = \frac{1}{N} \sum_{i=1}^{N}(\mathcal{L}_{cls}(f_{i}^{p})+\mathcal{L}_{tri}(f_{i}^{p}),
\end{split}
\end{equation}
where the $f^g$ and $f^p$ refer to the final output of global class token and part tokens, $\alpha$ and $\beta$ are hyper-parameters to balance the weight of attention diversity loss $\mathcal{L}_{adc}$ and correlation diversity loss $\mathcal{L}_{cdc}$, respectively. During the testing stage, we add the global class token $f^g$ with all the part tokens $f^p$ as:
\begin{equation}
f = \frac{1}{N+1}(f^g + \sum_{N}^{i=1}f_{i}^{p}).   
\end{equation}

\section{Experiments}
\label{sec:Experiments}
\subsection{Datasets and Experimental Setting}
\textbf{Datasets.} To evaluate the effectiveness of the proposed PartFormer, we conduct extensive experiments on six publicly available object Re-ID benchmarks, which include four person Re-ID and two vehicle Re-ID datasets. The details of these datasets are as follows. \textbf{Market-1501}~\cite{zheng2015scalable} is a widely-used Re-ID dataset captured from 6 cameras. It includes 12,936 images of 751 persons as the training set, 3,368 images of 750 persons as the query, and 19,732 images of 750 persons as the gallery. \textbf{DukeMTMC-reID}~\cite{zheng2017unlabeled} contains 36,441 images of 1,812 persons captured by eight cameras, in which 16,522 images of 702 identities are used as the training set, 2,228 and 16,522 images of 702 persons that do not appear in the training set are used as the query and gallery, respectively. \textbf{Occluded-Duke}~\cite{miao2019pose} is a dataset collected from the DukeMTMC for occluded person Re-ID. The training set consists of 15,618 images of 702 persons. The testing set contains 2,210 images of 519 persons as the query and 17,661 images of 1,110 persons as the gallery. \textbf{VeRi-776}~\cite{VeRi776} consists of 49,357 images of 776 distinct vehicles captured by 20 cameras in different orientations and lighting conditions. Among them, 576 identities (37,778 images) and 200 identities (11,579 images) are selected for training and testing respectively. Furthermore, 1,678 images from 200 identities are selected as the query from the testing set. \textbf{VehicleID Dataset}~\cite{VehicleID} is composed of 221,567 images from 26,328 unique vehicles. Half of the identities are used for training while the other half is for testing. There are 6 testing split strategies with various gallery sizes. Following the TransReID~\cite{he2021transreid}, we adopt the gallery size as 800.

\textbf{Evaluation Protocol.} To verify fair comparison with other methods, we adopt the widely used Cumulative Matching Characteristic (CMC) and mean Average Precision (\emph{m}AP) as evaluation metrics.

\textbf{Implementation details.}
We employ the ViT~\cite{dosovitskiy2020image} pre-trained on ImageNet~\cite{deng2009imagenet} as the backbone network and follow the training details with TransReID~\cite{he2021transreid} including the learning rate, decays at a cosine learning rate, batch size, etc. 
All the person images are resized to $256 \times 128$ and all vehicle images are resized to $256 \times 256$. Commonly used horizontal flipping, padding, random cropping, and random erasing~\cite{zhong2020random} are employed as data augmentation. We implement our PartFormer with PyTorch and conduct all experiments on the Nvidia A100.

\begin{table*}[t]
\centering
\scriptsize
\renewcommand{\arraystretch}{1.6}
\renewcommand{\tabcolsep}{3pt}
\resizebox{2.1\columnwidth}{!}{
\begin{tabular}{l|cccccccc||l|cccc}
 \hline
  \multicolumn{1}{c|}{} & \multicolumn{2}{c}{MSMT17} & \multicolumn{2}{c}{Market1501} & \multicolumn{2}{c}{DukeMTMC} & \multicolumn{2}{c||}{Occluded-Duke} & &\multicolumn{2}{c}{VeRi-776} &\multicolumn{2}{c}{VehicleID}\\
  Method & \emph{m}AP  & R1  & \emph{m}AP & R1  & \emph{m}AP  & R1  & \emph{m}AP  & R1 & Method     &  \emph{m}AP  & R1 & R1  & R5\\
 \hline
 \hline
 \multicolumn{5}{l}{\textbf{ResNet-50~\cite{he2016deep}}}\\
 \hline
    MGN~\cite{wang2018learning}     & 52.1  & 76.9 & 86.9 & 95.7 & 78.4 & 88.7 & - &- & UMTS~\cite{jin2020uncertainty}  & 75.9&95.8 & 80.9& 87.0 \\
    SCSN~\cite{SCSN}                & 58.5 & 83.8 & 88.5 & 95.7 & 79.0 & 91.0 & - &- & PGAN~\cite{PGAN}& 79.3 & 96.5 & 78.0 & 93.2\\
    ABDNet~\cite{ABD-Net}           & 60.8 & 82.3 & 88.3 & 95.6 & 78.6 & 89.0 & - &- & PVEN~\cite{PVEN} & 79.5&95.6 & 84.7& 97.0 \\
    AGW~\cite{ye2021deep}           & 49.3 & 68.3 & 87.8 & 95.1 & 79.6 & 89.0 & - & - & CFVMNet~\cite{sun2020cfvmnet}  & 77.1&95.3 & 81.4& 94.1 \\
    ISP~\cite{ISP}                  & - & - & 88.6 & 95.3 & 80.0 & 89.6 & 52.3 &62.8 & SN++~\cite{shen2021exploring} & 75.7 & 95.1 & 76.7 & 87.0\\
    PAT~\cite{li2021diverse}        & - & - & 88.0 & 95.4 & 78.2 & 88.8 & 53.6 & 64.5& HPGN~\cite{shen2021exploring} & 80.2 & 96.7 & 83.9 & -\\
    CAAO~\cite{zhao2023content}     & - & - & 87.3 & 95.1 & 77.5 & 88.9 & 55.8 & 67.8& MMNet~\cite{tu2022multi} & 80.0 & 95.6 & 82.1 & 95.3\\
    RFCNet~\cite{hou2022feature}    & 60.2 & 82.0 & 89.2 & 95.2 & 80.7 & 90.7 & 54.5 & 63.9& DFNet\cite{bai2022disentangled} & 81.0 & 97.1 & 84.8 & 96.2\\
    AutoLoss~\cite{gu2022autoloss}  & 63.0 & 83.7 & 90.1 & 96.2 & - & - & - & - & BIDA~\cite{Li2023bi} & 82.4 & 96.5 & 86.2 & 97.3\\
 \hline
 \multicolumn{5}{l}{\textbf{ViT-Base~\cite{dosovitskiy2020image}}}\\
 \hline
 DCAL~\cite{zhu2022dual} & 64.0 & 83.1 & 87.5  &94.7  &80.1 &89.0 &- &- & DCAL~\cite{zhu2022dual}  & 80.2 & 96.9 &- &-\\
    TransReID~\cite{he2021transreid} & 64.9  &83.3  &88.2  &95.0  &80.6 &89.6 &55.7 &64.2 & TransReID~\cite{he2021transreid} &80.6 &96.9 &83.6&97.1\\
    \textbf{PartFormer} & \textbf{68.3}  & \textbf{85.2}  & \textbf{89.7}  & \textbf{95.6}  & \textbf{82.4} & \textbf{91.1} & \textbf{58.8} & \textbf{66.3}   &\textbf{PartFormer} &\textbf{82.2} &\textbf{97.5} &\textbf{84.9}&\textbf{97.4}\\
    \hline
    TransReID*~\cite{he2021transreid} & 67.4 & 85.3 & 88.9 & 95.2 &82.0 &90.7 &59.2 &66.4 & GiT*~\cite{shen2023git} & 80.3  & 96.9  & 84.7  & -\\
    PFD*~\cite{wang2022pose} & 64.4 & 83.8 & 89.7 & 95.5 &83.1 & 91.2 &61.8 &69.5 & MsKAT$\dag$~\cite{li2022mskat} & 82.0  & 97.1  &86.3  & 97.4\\
    PHA*~\cite{zhang2023pha} & 68.9 & 86.1 & 90.2 & 96.1 &- &- &- &- & TransReID*~\cite{he2021transreid} &82.0  & 97.1  &85.2  & 97.5\\
    \textbf{PartFormer}* & \textbf{71.3}  & \textbf{87.1}  &\textbf{90.7}  &\textbf{96.1}  & \textbf{83.6} &\textbf{91.6} & \textbf{62.5} & \textbf{70.2} &\textbf{PartFormer}* &\textbf{82.9} &\textbf{97.7} &\textbf{86.6}&\textbf{98.0}\\
 \hline
\end{tabular}}
\vspace{0.5em}
\caption{\label{tab:sota}\textbf{Comparision with state-of-the-art methods in terms of CMC (\%) and \emph{m}AP (\%) on person Re-ID (left) and vehicle Re-ID datasets (right).} The * means the backbone is with a sliding window setting. The $\dag$ indicate that the methods using both CNN and ViT.}
\end{table*}

\subsection{Comparison with State-of-the-art Methods}
To comprehensively evaluate the performance of PartFormer, we compared it against previously reported state-of-the-art methods on both person and vehicle Re-ID datasets, as shown in Table~\ref{tab:sota}. Given that TransReID employs a sliding window strategy in its patchify process, which is broadly adopted in later works, we present results under both settings. On person Re-ID tasks, PartFormer achieves a marked improvement over the previous methods in handling both holistic (MSMT17, Market1501, and DukeMTMC) and occluded scenarios (Occluded-Duke), as evidenced in the Rank-1 and \emph{m}AP metrics. Specifically, compared to TransReID and PFD, which similarly explore fine-grained representations within ViT architectures, PartFormer surpasses them with at least a $3.9\%$ and $1.8\%$ margin in Rank-1 and \emph{m}AP metrics on the largest person Re-ID dataset, MSMT17. This success extends to vehicle Re-ID, where PartFormer achieves impressive results on both the VeRi-776 and VehicleID datasets. Compared to DCAL, GiT, and TransReID, which also utilize a pure Vision Transformer architecture, PartFormer exhibits an improvement of at least $0.7\%$ in Rank-1 and $0.4\%$ in \emph{m}AP on the VeRi-776 dataset, and $1.4\%$ and $0.5\%$ in Rank-1 and Rank-5, respectively, on VehicleID.

\begin{table}[t]
	\centering
	\footnotesize
    \renewcommand{\arraystretch}{1.4}
    \resizebox{0.95\columnwidth}{!}{
	\begin{tabular}{ll|cc|cc}
		\hline
		\multicolumn{2}{c|}{\multirow{2}{*}{Settings}} &\multicolumn{2}{c|}{MSMT17} & \multicolumn{2}{c}{VeRi-776} \\
		& & \emph{m}AP & R1  & \emph{m}AP & R1 \\
		\hline
		\hline
		(a) & B               & 62.6  & 82.1  & 79.7  & 97.0\\
        (b) & B(head=12)      & 62.4  & 81.9  & 79.6  & 96.9\\
		\hline
		(c) & + HDB         & 65.9 &  83.5 &  80.8 & 97.2 \\
		(d) & + ADC         & 63.5 &  82.7  & 80.3 & 97.1 \\
		\hline
		(e) & + HDB + ADC & 67.2 & 84.4 & 81.3 & 97.3 \\
		(g) & + HDB + ADC + CDC & \textbf{68.3} & \textbf{85.2} & \textbf{82.2} & \textbf{97.5} \\
		\hline
	\end{tabular}}
	\vspace{0.5em}\caption{\textbf{Ablation analyses of PartFormer.} ``B'' denotes the baseline of our work which is based on a vanilla ViT. ``HDB'' refers to the Head Disentangling Block. ``ADC'' and ``CDC'' indicate the Attention Diversity Constraint and Correlation Diversity Constraint. As can be observed, each proposed component plays an important role and contributes to the performance.}
	\label{tab:ablation}
\end{table}

\begin{figure}[t]
    \centering
    \includegraphics[width=1.0\columnwidth]{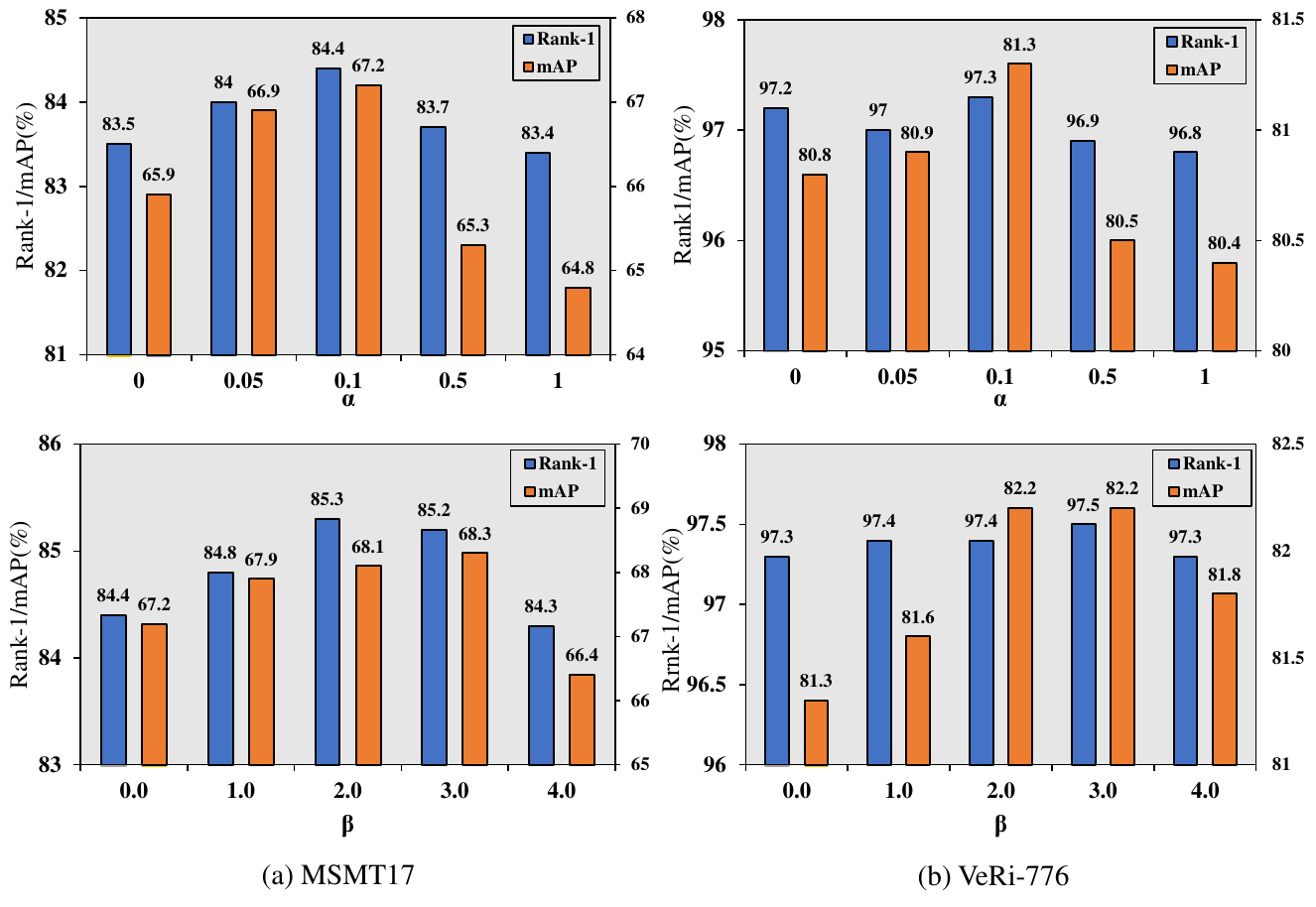}
    \caption{\textbf{Analysis of parameters $\alpha$ and $\beta$ in Eq.~(\ref{eq:loss}).} The optimal performance reaches when $\alpha = 0.1$ and $\beta = 3.0$.}
    \label{fig:hyper}
\end{figure}

\subsection{Ablation Study}
To evaluate the impact of the different components, we conducted a series of experiments on MSMT17 and VeRi-776 using varying configurations, with the quantitative results presented in Table~\ref{tab:ablation}. In PartFormer, we reduced the number of heads in the HDB from 12 to 6. Consequently, we provide comparisons with the original baseline configuration, where the number of heads is set to 12.

Replacing the final transformer block with the HDB resulted in improvements of 3.3\% in \emph{m}AP and 1.4\% in Rank-1 on the MSMT17, as well as gains of 1.1\% in \emph{m}AP and 0.2\% in Rank-1 on the VeRi-776. These performance gains validate the effectiveness of the HDB in exploiting the latent diversity within the MHSA mechanism. Performance is further enhanced when combining the attention diversity constraint (ADC), as the ADC imposes an explicit constraint on the focus regions of different heads. Furthermore, incorporating the ADC into the baseline model proves beneficial, indicating that maintaining diversity among heads is critical in the Vision Transformer, even with head selective processing. Additionally, the correlation diversity constraint (CDC), designed to augment content diversity across different heads, also significantly enhances PartFormer's performance. Deploying the CDC yields a $1.1\%$ rise in Rank-1 and a $0.8\%$ uplift in \emph{m}AP on the MSMT17, alongside a $0.9\%$ increase in Rank-1 and a $0.1\%$ advance in \emph{m}AP on VeRi-776. Consequently, each component contributes consistently to the effectiveness of the framework, culminating in PartFormer's impressive performance.

\subsection{Discriminative Evaluations}
\begin{table}[t]
	\centering
	\footnotesize
    \renewcommand{\arraystretch}{1.4}
	\begin{tabular}{ll|c|cc|cc}
		\hline
		\multicolumn{2}{c|}{\multirow{2}{*}{Settings}}& \multirow{2}{*}{$N$} &\multicolumn{2}{c|}{MSMT17} & \multicolumn{2}{c}{Veri-776} \\
		& & & \emph{m}AP & R1  & \emph{m}AP & R1 \\
		\hline
		\hline
        (a) & PartFormer       &   2    & 64.2  & 62.5  & 81.0  & 97.1\\
		(b) & PartFormer       &   3    & 67.1  & 84.9  & 81.8  & 97.2\\
        (c) & PartFormer       &   6    & \textbf{68.3}  & \textbf{85.2}  & \textbf{82.2}  & 97.5\\
        (d) & PartFormer       &   12   & 67.8  & 85.0  & 81.6  & \textbf{97.7}\\
		\hline
	\end{tabular}
	\vspace{0.5em}\caption{\textbf{Performance under different numbers of heads (N) in HDB.} The optimal reaches when $N$ is set to $6$.}
	\label{tab:head}
\end{table}

\begin{table}[t]
	\centering
	\footnotesize
    \renewcommand{\arraystretch}{1.4}
	\setlength{\tabcolsep}{6.5pt}
    \resizebox{\columnwidth}{!}{
	\begin{tabular}{ll|cc|cc}
		\hline
		\multicolumn{2}{c|}{\multirow{2}{*}{Settings}} &\multicolumn{2}{c|}{MSMT17} & \multicolumn{2}{c}{Occluded-Duke} \\
		& & \emph{m}AP & R1  & \emph{m}AP & R1 \\
		\hline
		\hline
        (a) & ViT                        & 62.4  & 81.9  & 53.2  & 60.4\\
		(b) & Horizontal Partition            & 60.4  & 79.3  & 54.2  & 61.7\\
        (c) & Patch Merging         & 63.5  & 82.3  & 54.9  & 62.6\\
        (d) & ViT-PAT & 63.6  & 81.8  & 55.5  & 63.2\\
	  \hline
		(d) & PartFormer & \textbf{68.3} & \textbf{85.2} & \textbf{58.8} & \textbf{66.3} \\
		\hline
	\end{tabular}}
	\vspace{0.5em}\caption{\textbf{Performance under different partition strategies.} Compared to different partition approaches, PartFormer shows a significant advantage.}
	\label{tab:partition}
\end{table}

\textbf{Impact of the hyper-parameters.}
PartFormer contains several hyperparameters, such as the $\alpha$ and $\beta$ in Eq.~(\ref{eq:loss}), as well as the number of heads $N$ in the HDB. Therefore, in this section, we first analyze the influence of the $\alpha$ and $\beta$ on the performance of PartFormer in Fig.~\ref{fig:hyper}. On the MSMT17 dataset, we observe a linear improvement in performance when $\alpha$ is less than $0.1$. Beyond this point, further increasing the $\alpha$ will decrease performance. For the VeRi dataset, while Rank-1 exhibits fluctuations, optimal performance is still achieved when $\alpha$ is set to $0.1$. Regarding $\beta$, performance consistently improves when $\beta$ is less than $2.0$. Beyond this, comparable peak performance is observed at both $\beta = 2.0$ and $\beta = 3.0$. We finally set the $\beta$ to $3.0$ in further experiments due to its slight advantage in VeRi-776. Additionally, we investigate the influence of the number of heads $N$, within the HDB and present the results in Table~\ref{tab:head}. Due to the limitation of the feature dimension, $N$ can only be set to several specific values. The data indicate that when $N$ is set to 6 yields the best results, with any deviation from this number leading to reduced performance across both datasets.

\textbf{Comparison between different partition strategies.} 
Adopting a uniform partition or prototype-based feature factorization to divide the person image has been an influential method for achieving fine-grained representations within CNN models~\cite{sun2018beyond,wang2018learning}. Therefore, we compare the head-based soft partition strategy of PartFormer with two distinct hard partition strategies (Horizontal Partition and Patch Merging Partition) and a prototype-based feature factorization (PAT~\cite{li2021diverse}). As detailed in Table~\ref{tab:partition}, for a fair comparison, image patches are divided into $N$ parts within the final transformer block to compute their feature representations independently. Due to the different designs between CNN and ViT, which have different computation strategies to obtain the final feature map, these partition strategies work well in CNN but are not so efficient in ViT, as observed in the MSMT17. This suggests that directly applying fine-grained strategies that are effective in CNNs may not be suitable for ViTs.
Additionally, drawing inspiration from the Patch Merging feature of Swin-Transformers~\cite{liu2021swin}, we implemented a Patch Merging Partition that utilizes the Patch Merging operation to partition image patches into $N$ segments. While the Patch Merging Partition is viable for both person Re-ID datasets, it still falls significantly short of PartFormer's performance.

\begin{table}[t]
\centering
\footnotesize
\renewcommand{\arraystretch}{1.5}
\renewcommand{\tabcolsep}{3pt}
\begin{tabular}{r|c|c|c|c}
\hline
 Method & ViT-Base & TransReID & PFD & PartFormer\\
 \hline
 \hline
 MSMT17         & 62.4/81.9  &  64.9/83.3  & 64.4/83.8 & 68.3/85.2\\
 VeRi-776        & 79.6/96.9  & 80.6/96.9  & -/- & 82.2/97.5\\
 Training Time   & 1.0$\times$ & 1.5$\times$ & 6.6$\times$ & 1.3$\times$ \\
 GPU Memory            & 1.0$\times$ & 1.1$\times$ & 1.3$\times$ & 1.1$\times$  \\
 \hline
\end{tabular}
\vspace{0.5em}\caption{\textbf{Consumption comparison with other part-based strategies.} Here, the performance is shown as \emph{m}AP/Rank-1.}
\label{tab:computation}
\end{table}

\textbf{Computation Consumption Comparison.} 
To evaluate the efficiency of PartFormer, we compare it with two part-based ViT models~\cite{he2021transreid,wang2022pose} in terms of training time and GPU consumption. A vanilla ViT (ViT-Base) is considered the baseline model. Compared to the vanilla ViT, due to the addition of part branches, the computation cost increased in PartFormer. However, compared to other part-based ViT models, PartFormer shows a strong advantage when considering both performance and training consumption.

\begin{table}[t]
	\centering
	\footnotesize
    \renewcommand{\arraystretch}{1.4}
	\setlength{\tabcolsep}{6.5pt}
    \resizebox{\columnwidth}{!}{
	\begin{tabular}{ll|cc|cc}
		\hline
		\multicolumn{2}{c|}{\multirow{2}{*}{Settings}} &\multicolumn{2}{c|}{MSMT17} & \multicolumn{2}{c}{Veri-776} \\
		& & \emph{m}AP & R1  & \emph{m}AP & R1 \\
		\hline
		\hline
        (a) & w/ Residual        & 62.4  & 81.4  & 81.7  & \textbf{97.6}\\
        (b) & w/ FFN             & 66.7  & 84.1  & 81.8  & 97.3\\
        (c) &w/o cls             & 63.2  & 82.8  & 81.9  & 97.3\\
	  \hline
		(d) & PartFormer & \textbf{68.3} & \textbf{85.2} & \textbf{82.2} & 97.5 \\
		\hline
	\end{tabular}}
	\vspace{0.5em}\caption{\textbf{Performance under different settings of HDB.} "w/ Residual'' denotes adding the residual connect in the HDB. ``w/ FFN'' refers to adding the FFN layers in the HDB block. ``w/o cls'' indicates the removing the class token in HDB.}
	\label{tab:hdbset}
\end{table}

\textbf{Dissusion on Different Setting of PartFormer.} 
In PartFormer, we made several modifications to the HDB to gain better performance. To evaluate the influence of these modifications, we conducted several experiments and showed the results in Table.\ref{tab:hdbset}. To limit the influence from the previous blocks, we remove the residual connect in the HDB, which shows an improvement in nearly all the settings. Additionally, without the necessity to further select the feature from multi-heads, we remove the FFN in the HDB. This not only reduced the computation of PartFormer but also yielded performance gains. Finally, we attempt to remove the class token in the HDB and train the PartFormer with the part tokens. Unfortunately, the class token proved to be important in guiding the training of heads. Its removal led to a significant decline in performance.
\begin{figure}
    \centering
    \includegraphics[width=\linewidth]{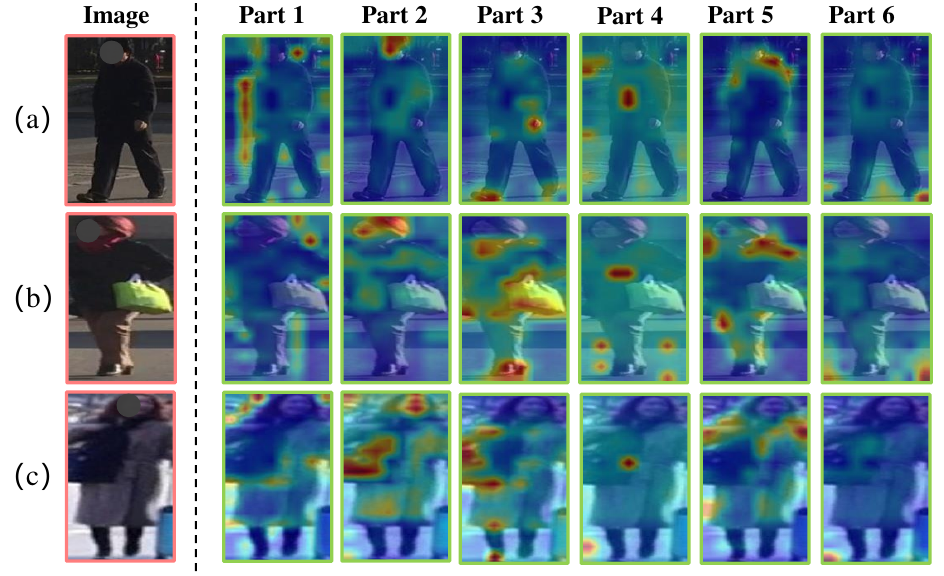}
    \caption{Visualization results of PartFormer on the MSMT17.}
    \label{fig:attvis}
\end{figure}

\textbf{Visualization results.} 
In Fig.~\ref{fig:attvis}, we visualized the attention results of different part tokens. From the results, we can observe that different part tokens can successfully capture diverse representations from different human parts for the same input image. Specifically, Part 1 mainly focuses on the contour of the person, Part 2 mainly focuses on the head, Part 3 mainly focuses on the hands and shoes, Part 4 mainly focuses on the center part of the person, and Part 5 mainly focuses on the shoulder. There may be some differences in Part 6, which mainly focuses on the corner of the image. Due to the similarity-based computation strategy in ViT, the corner without person content will show uniform-like attention results. Therefore, we can consider the representation of Part 6 as an average pooling of all the image patches.

\subsection{Limitations}
The primary limitation of Partformer is that although Partformer aims to awaken the diversity representation within the ViT. Owing to structural constraints, we are limited to substituting the final block of the ViT with HDB. Incorporating additional HDB layers would lead to an exponential growth in the count of tokens, thereby rendering the computationally infeasible. This condition largely limited the effectiveness of PartFormer. Nonetheless, we still believe that the ability of Partformer still has a large room to improve, which will be left in our future work.

\section{Conclusion}
\label{sec:Conclusion}
In this paper, we propose a novel part-based vision transformer, termed PartFormer, for object Re-ID. PartFormer aims to explore a part-based, fine-grained representation within a vision transformer. By observing the latent diverse representation behind the multi-head attention, we introduce a Head Disentangling Block to awaken the latent diverse representation. Moreover, to prevent heads’ representation from inevitably homogenizing, we incorporate two diversity constraints, including an attention diversity constraint and a correlation diversity constraint to explicitly encourage different heads to show a diverse representation. Extensive experiments across various object Re-ID benchmarks have proved the necessity of diverse part-based representation in PartFormer and demonstrated its state-of-the-art performance.

\section*{Acknowledgement}
This work was supported by National Science and Technology Major Project (No. 2022ZD0118202), the National Science Fund for Distinguished Young Scholars (No.62025603), the National Natural Science Foundation of China (No. U21B2037, No. U22B2051, No. U23A20383, No. 62176222, No. 62176223, No. 62176226, No. 62072386, No. 62072387, No. 62072389, No. 62002305 and No. 62272401), and the Natural Science Foundation of Fujian Province of China (No.2022J06001).










\bibliography{sn-bibliography}


\end{document}